\documentclass[conference]{IEEEtran}
\IEEEoverridecommandlockouts
\usepackage{cite}
\usepackage{amsmath,amssymb,amsfonts}
\usepackage{graphicx}
\usepackage{textcomp}
\usepackage{xcolor}
\usepackage{balance}
\usepackage{multirow}
\usepackage{url}
\usepackage{makecell}
\usepackage{amsthm}
\newtheorem{example}{Example}
\newtheorem{definition}{Definition}
\usepackage{enumitem}
\usepackage[ruled, lined, vlined]{algorithm2e}
\usepackage{amsmath}
\usepackage{ulem}

\begin{document}

\title{SAGDFN: A Scalable Adaptive Graph Diffusion Forecasting Network for Multivariate Time Series Forecasting }
\author{\IEEEauthorblockN{Yue Jiang$^{1,4}$, Xiucheng Li$^{2}$\IEEEauthorrefmark{1}, Yile Chen$^{1}$, Shuai Liu$^{1}$, Weilong Kong$^{1}$, Antonis F. Lentzakis$^{3}$, Gao Cong$^{1}$\IEEEauthorrefmark{1}}
\IEEEauthorblockA{$^1$Nanyang Technological University, Singapore \\
$^2$Harbin Institute of Technology(Shenzhen), China\\
$^3$NCS Pte Ltd, Singapore\\
$^4$DAMO Academy, Alibaba group, Singapore\\
\{yue013@e., yile001@e., shuai004@e., weilong.kong@, gaocong@\}ntu.edu.sg\\
lixiucheng@hit.edu.cn,
antonis.lentzakis@ncs.com.sg}
\thanks{\IEEEauthorrefmark{1} Corresponding Author.}
}

\maketitle

\begin{abstract}
Time series forecasting is essential for our daily activities and precise modeling of the complex correlations and shared patterns among multiple time series is essential for improving forecasting performance. Spatial-Temporal Graph Neural Networks (STGNNs) are widely used in multivariate time series forecasting tasks and have achieved promising performance on multiple real-world datasets for their ability to model the underlying complex spatial and temporal dependencies.
However, existing studies have mainly focused on datasets comprising only a few hundred sensors due to the heavy computational cost and memory cost of spatial-temporal GNNs. When applied to larger datasets, these methods fail to capture the underlying complex spatial dependencies and exhibit limited scalability and performance. To this end, we present a \textbf{S}calable \textbf{A}daptive \textbf{G}raph \textbf{D}iffusion \textbf{F}orecasting \textbf{N}etwork (SAGDFN) to capture complex spatial-temporal correlation for large-scale multivariate time series and thereby, leading to exceptional performance in multivariate time series forecasting tasks. The proposed SAGDFN is scalable to datasets of thousands of nodes without the need of prior knowledge of spatial correlation. 
Extensive experiments demonstrate that SAGDFN achieves comparable performance with state-of-the-art baselines on one real-world dataset of 207 nodes and outperforms all state-of-the-art baselines by a significant margin on three real-world datasets of 2000 nodes.
\end{abstract}

\section{Introduction}
Multivariate time series data records the quantities of interest evolving over time and arises in many aspects of scientific research and human activities such as meteorology, stock market, traffic flow, energy consumption, etc., and thus the accurate multivariate time series forecasting plays a central role in decision-making, investment management, trip arrangement, and production schedule. 

There are two key factors in describing the dynamics of multivariate time series, namely, temporal dependency and spatial dependency (correlation). The temporal dependency characterizes the dynamics of time series along the time dimension and the spatial dependency specifies how different series are correlated with each other along the instance (series) dimension. To develop an accurate multivariate time series forecasting model, it is equally important to faithfully model the dynamics of time series from both dimensions. However, some methods, including the statistical models, such as ARIMA, VAR~\cite{ARIMA, VAR}, machine learning models, such as SVR, and Gaussian Process~\cite{SVR, gaussianprocess}, and the RNNs- and Transformer-based forecasting methods\cite{GRU, LSTM, informer, Autoformer}, mainly focus on developing models to describe temporal dynamics.
The proposals~\cite{DCRNN, TGCN} apply GNNs with RNNs jointly to forecast the traffic flow. 
However, they are only applicable to the scenarios where the graph topology is available (e.g., road network) or can be explicitly constructed (constructing the graph according to spatial proximity), which brings two main limitations. First, the graph topology is not always clearly defined, e.g., the trend of stock price data; second, the graph topology induced by the spatial proximity may not reflect the true correlations, i.e., two nearby time series can be uncorrelated and two distant ones are likely to exhibit similar trend due to latent common causes ~\cite{AGCRN, GRAPHWaveNet, MTGNN}.

To address these two limitations, the adaptive-weight-GNN forecasting methods~\cite{AGCRN, MTGNN, GRAPHWaveNet, GTS, STEP, D2STGNN, STGCN} are proposed. The key idea is to learn the adjacency matrix or weight matrix directly from data and integrate it with the forecasting model, under the guidance that an optimal adjacency matrix or weight matrix should provide the true correlations and aid in the forecasting. Specifically, the adaptive-weight-GNN methods learn each node $i$ an embedding $\mathbf{e}_i$ 
and compute the pairwise correlation $A_{ij}$ between node $i$ and node $j$ as $A_{ij} = f(\mathbf{e}_i, \mathbf{e}_j)$ where $f(\cdot, \cdot)$ is a transformation function such as inner product or feed-forward neural networks. Several studies~\cite{ASTGCN, ASTGNN, DGCN, ST-WA} leverage the attention mechanism to learn the underlying spatial correlations. For instance, ASTGNN~\cite{ASTGNN} proposes a temporal-aware attention-based network to enhance forecasting ability. Similarly, ST-WA~\cite{ST-WA} presents a spatio-temporal aware network, utilizing a window attention technique to mitigate temporal complexity.
The adaptive-weight-GNN methods show significant advancements over the prior GNN-based methods with predefined graph structures. Despite their performance superiority, the adaptive-weight-GNN forecasting methods still have two main limitations.

First, they require computing and utilizing an $N \times N$ weight matrix, which is quite expensive in terms of both computation complexity and memory complexity, and thus fail to scale to large datasets. Table~\ref{tab:complexity} shows the complexity of three representative adaptive-weight-GNN forecasting methods, with both their computational and memory costs growing quadratically with the number of nodes or time series $N$. It becomes prohibitively costly to apply them to relatively large multivariate time series forecasting datasets, e.g., all of them will run into out-of-memory errors on a 32GB memory GPU when $N$ grows out of 2000.
Take GTS \cite{GTS} as an example, we show its memory consumption analysis in Example~\ref{firstexample}. 
\begin{example}\label{firstexample}
First, such a typical multivariate time series forecasting setup leads to the model's hidden state variables of size $B \times N \times T \times D$, resulting in $64\times 2000 \times 64\times 24 \times 8 Byte = 1.57 GB$ GPU memory per variable. 
Second, the CUDA memory runs out quickly in order to store the node embeddings of size $N\times N\times d$ (d is the embedding dimension).
Last, given the computation $B\cdot T\cdot D \cdot N = 64\cdot24\cdot64\cdot N = 98,304N$, GPU memory consumption escalates by almost 100,000 times proportional to the graph size.
As the graph size grows to 2000 nodes, storing all intermediate states and the node embeddings costs substantial GPU resources and underscores the forecasting performance~\cite{GNNreally, MTGNN}.
\end{example}
In practice, as we will show in Section~\ref{sec:experiments}, a larger $N$ could potentially offer more structure information.
However, the adaptive-weight-GNN forecasting methods all fail to explore it due to their exhausting memory usage. Several methods~\cite{graphsaint, clustergcn, scaleup, Shadow, Nystrom} such as GraphSAINT~\cite{graphsaint}, Cluster-GCN~\cite{clustergcn}, and SHADOW~\cite{Shadow} have been developed to tackle the computational expense associated with traditional Graph Neural Networks (GNNs). These strategies implement various sampling, sub-graph partitioning or coarsening, and graph clustering algorithms to mitigate computational costs. For instance, the Nystrom method~\cite{Nystrom} provides an effective spectral clustering solution by approximating global eigenvectors using a locally sampled graph. However, these approaches do not align smoothly with our research context. This is due to their inherent reliance on a known adjacency matrix.

Second, such a manner fails to take spatial sparsity into consideration and can bring noise to the graph convolution operation. Intuitively, a node (time series) is often only correlated with a small fraction of nodes in real-world scenarios, i.e., its neighbor size should be small. However, the existing adaptive-weight-GNN methods implicitly assume a dense connection for the underlying graph, and $A_{ij}$ will be a trivial but nonzero value even if node $i$ and $j$ have no correlation. This may not affect the forecasting performance on small graphs; but when $N$ becomes large the accumulated noise entries can be significant and impede the forecasting performance.

\begin{table}
    \centering  
    \caption{Complexity of adaptive-weight-GNN forecasting methods ($N$, $d$, $D$, $M$ are \# of nodes, node embedding dimension, hidden dimension, and size of the most significant nodes, respectively).}
    \label{tab:complexity}
    \begin{tabular}{c|c|c}  
         \hline 
         Model& Computation Complexity & Memory Complexity\\
         \hline 
         AGCRN & $\mathcal{O}(N^2d+N^2D)$& $\mathcal{O}(N^2+Nd)$\\
         GTS  & $\mathcal{O}(N^2d^2+N^2D)$& $\mathcal{O}(N^2+N^2d)$\\
         STEP & $\mathcal{O}(N^2d^2+N^2D)$& $\mathcal{O}(N^2+N^2d)$\\
         SAGDFN & $\mathcal{O}(NMd^2+NMD)$& $\mathcal{O}(NM +NMd)$\\
         \hline
    \end{tabular}
 \vspace{-0.45cm}
\end{table}

To address these two limitations, we propose a Scalable Adaptive Graph Diffusion Forecasting Network. Our key idea is to take advantage of the spatial sparsity to simultaneously reduce the complexity and the negative effect caused by the noise entries. Specifically, we propose to select the most significant $M$ ($M \ll N$) nodes that are globally influential to all $N$ nodes in an end-to-end learning approach. This is achieved by a simple but effective Significant Nodes Sampling algorithm, and the output of the sampling algorithm is fed to a well-developed Sparse Spatial Multi-Head Attention module to refine and produce a slim adjacency matrix $\mathbf{A}_s$ with size $N \times M$, and the $\alpha$-Entmax function is adopted to further encourage sparsity. The slim matrix $\mathbf{A}_s$ is then used to perform graph diffusion. Both the computational and memory cost are reduced from $N^2$ to $MN$ as shown in Table \ref{tab:complexity}.
\begin{example}
In comparison to Example \ref{firstexample}, the GPU consumption in graph diffusion is reduced to $B \times M \times T \times D$, typically amounting to less than 0.1GB per hidden state variable. Additionally, the GPU consumption for node embeddings is reduced from 64GB to 3.2GB, achieved by setting $M$ to 100.
\end{example}
Our idea also mitigates the negative effects induced by the noise entries. This is because only the most significant nodes are considered and refined by the $\alpha$-Entmax function. Moreover, the Significant Nodes Sampling algorithm and Sparse Spatial Multi-Head Attention module interact with the subsequent forecasting module in a differentiable manner, and thus all parameters can be learned end-to-end. In summary, our contributions are as follows.  
\begin{itemize}
\item We propose a scalable adaptive graph diffusion network for the multivariate time series forecasting problem. It is able to scale to large datasets and exploit the spatial correlation in an efficient fashion.
\item We introduce a Significant Neighbor Sampling algorithm and a Sparse Spatial Multi-Head Attention module, by which we can derive a slim dense adjacency matrix that enables reducing the complexity from $\mathcal{O}(N^2)$ to $\mathcal{O}(MN)$, where $M \ll N$, in addition to mitigating the negative effect caused by noise entries.
\item Extensive experiments on four real-world datasets demonstrate that our proposed model not only attains state-of-the-art performance on widely used datasets but also outperforms other state-of-the-art models on three large datasets. The majority of the state-of-the-art models fail to handle these three large datasets.
\end{itemize}

\section{RELATED WORK}\label{sec:relatedwork}
\subsection{Statistical Models and Machine Learning Methods}
In earlier studies, multiple classical statistical models are used to model the time series forecasting problems \cite{ARIMA, VAR}. For example, ARIMA employs autoregression, differencing, and moving averages techniques to model the linear relationships of the time series. Vector autoregression (VAR) \cite{VAR} models multiple time series together in an autoregression approach using maximum likelihood estimation. The classical machine learning models such as Support Vector Regression (SVR) \cite{SVR} tries to find a hyperplane that maximizes the margin between the historical inputs and the prediction values. However, these classical methods fail to capture the non-linearity in time series and lack complex pattern learning capacity, thus having limited ability to model the complex correlations.
\subsection{Classical Neural Network-based Methods}
Neural network-based methods have been widely used to address time series forecasting problems. Recurrent Neural Networks (RNNs) are commonly used to model time series data and Long Short-Term Memory (LSTM) and Gated Recurrent Units (GRUs) are two variants of the RNNs architecture. Furthermore, some researchers divide the time series data into multiple grids based on topological information and use Convolutional Neural Networks (CNNs) to model the spatial correlations \cite{CLTFP, ConvLSTM} and employs RNNs to model the temporal correlations. For instance, ST-ResNet is proposed by \cite{ST-ResNet} to predict the citywide crowd flows. A deep multi-view spatial-temporal network proposed by \cite{DMVST-Net} integrates CNNs and LSTM for taxi demand prediction. These methods are able to capture the non-linearity in time series data and demonstrate good forecasting performance. However, only grid-based spatial correlations are considered in these methods.

Transformer-based models~\cite{informer, Triformer, Autoformer, Pyraformer, Fedformer, Crossformer} have demonstrated significant promise in long-sequence forecasting due to their capacity to effectively capture temporal dependencies. For instance, Informer~\cite{informer} employs the sparsity of attention scores via KL divergence estimation, introducing the concept of Prob-Sparse self-attention. This approach allows for the complexity of $\mathcal{O}(h(\log h))$, where $h$ denotes the sequence length. Triformer~\cite{Triformer} introduces a novel approach of triangular, variable-specific patch attention, enhancing accuracy while maintaining a linear complexity of $\mathcal{O}(h)$. However, these models concentrate on reducing the complexity inherent in time domain. They may inadvertently neglect the vital spatial dependencies that exist between time series. Such neglect can considerably influence the effectiveness of forecasting in multivariate time series.
\subsection{Spatial-Temporal Graph Neural Networks}
Spatial-Temporal Graph Neural Networks (STGNNs) have been extensively utilized as a default framework to model the correlations among the time series~\cite{Wu_2021, MTGNN, DCRNN, AGCRN, GTS, TGCN, STGCN, GCRN, D2STGNN, STEP, ASTGCN, ASTGNN, DGCRN}. In the STGNNs framework, GNNs are utilized to model the intricate spatial relationships and to facilitate the message-passing between time series. 
Graph Convolutional Networks (GCNs) ~\cite{GCN} are widely used GNNs methods to process spatial dependency. 
Furthermore,
different variants of RNNs and Temporal convolutional networks (TCNs) are frequently used by STGNNs-based methods to model temporal dependency. 
STGNNs have demonstrated state-of-the-art performance on numerous forecasting datasets, by jointly modeling spatial and temporal correlation in an end-to-end approach. 

DCRNN \cite{DCRNN} employs GRUs to model time dependencies, and uses GNNs to model spatial correlations with a predefined adjacency matrix obtained from topology information. T-GCN \cite{TGCN} also approximates the adjacency matrix from topology information and captures spatial correlation with the classical GCNs. Since the adjacency matrix derived from the topology information is static, the forecasting performance of these models is prone to inaccuracies when there is imprecision in the prior knowledge.

Models \cite{AGCRN, MTGNN, D2STGNN, GTS, STEP} incorporate the adaptive adjacency matrix into the STGNNs. AGCRN \cite{AGCRN} learns a set of node embeddings, which are numerical representations of each node in a given graph. Specifically, the inner product of these embeddings is computed to learn the adaptive adjacency matrix utilized in GCNs. This adjacency matrix updating process allows GCNs to effectively capture the underlying structure and relationships present in graph data, ultimately enhancing forecasting performance. MTGNN \cite{MTGNN} is an innovative approach that incorporates GCNs into the Mix-hop Propagation Layer alongside dilated temporal convolution modules to model the temporal dependency. To effectively capture the spatial dependency, MTGNN learns both source and destination node embeddings, and replaces the GCNs adjacency matrix with the inner product of the bi-directional node embeddings. Although these models exploit the potential spatial correlation, they fail to reveal the nonlinear spatial correlation, and their ability to approximate spatial correlation is limited in terms of expressiveness when the graph becomes large.

Several methods employ a more explicit graph learning approach by utilizing feed-forward neural networks to model the pair-wise spatial correlation, as opposed to simply calculating the inner product of node embeddings. Unlike the aforementioned models such as AGCRN\cite{AGCRN} and MTGNN\cite{MTGNN}, GTS \cite{GTS} and STEP \cite{STEP} use the full training sequence as initial node embeddings.
Such node embeddings are concatenated node-to-node for computing pair-wise correlation weights by a graph learning block. GTS \cite{GTS} and STEP \cite{STEP} have achieved state-of-the-art performance on multiple multivariate time series forecasting datasets. 

While these STGNNs methods have demonstrated significant advancements in forecasting widely used datasets, their ability to effectively leverage the underlying spatial-temporal correlation and extend to larger datasets remains limited. To tackle this challenge and improve the forecasting performance, we propose the SAGDFN method, which captures the precise spatial dependency between nodes without depending on the predefined adjacency matrix that can lead to flawed results. We incorporate a Significant Neighbors Sampling module to diminish computational overhead, thereby enabling the method's application to extensive datasets consisting of thousands of nodes. Additionally, we introduce a Sparse Spatial Multi-Head Attention module to accurately reduce noise and utilize spatial correlation. 

\section{Preliminary}\label{sec:Preliminary}
We first give the definitions of multivariate time series data and multivariate network in our proposed method, then we formally present the multivariate time series forecasting problem.
\begin{definition} \textbf{Multivariate Time Series}.
    A multivariate time series records the quantities of interest generated by $N$ instances over $T$ time steps at a fixed interval (e.g., 5 minutes), and each observation is a $C$ dimension variable, where $C$ is referred to as the channel. The observations made by all $N$ instances at time step $t$ are denoted by $\mathbf{X}_t \in \mathbb{R}^{N \times C}$.
\end{definition}
\begin{definition}
\textbf{Multivariate Network}. A multivariate network can be represented as a graph $G = (V, E, A)$, where $V$ denotes the nodes of the graph with the size $N$  (e.g., number of time series) and $E$ denotes the set of edges, and $\mathbf{A} \in {\mathbb{R}^{N \times N}}$ denotes the adjacency matrix. For instance, in a traffic speed dataset, $V$ constitutes the set of traffic sensors (each node corresponds to a time series recorded by the sensor), and $E$ represents the set of dependencies between nodes.
\end{definition}
\begin{definition}
\textbf{Multivariate Time Series Forecasting}. 
The multivariate time series forecasting problem attempts to predict the future $f$ steps of multivariate time series starting from time step $t$, $\mathcal{X}_f = [\mathbf{X}_{t+1}$, $\mathbf{X}_{t+2}$, $\dots$, $\mathbf{X}_{t+f}] \in \mathbb{R}^{f \times N \times C}$, given the past $h$ steps historical observations, $\mathcal{X}_h = [\mathbf{X}_{t-h+1}$, $\mathbf{X}_{t-h+2}$, $\dots$, $\mathbf{X}_{t}] \in \mathbb{R}^{h \times N \times C}$, as well as the corresponding covariates (e.g., time of the day, day of the week).
\end{definition}
\section{SAGDFN FRAMEWORK}\label{sec:method}
\begin{figure*}[ht]
  \centering
  \includegraphics[width=\linewidth]{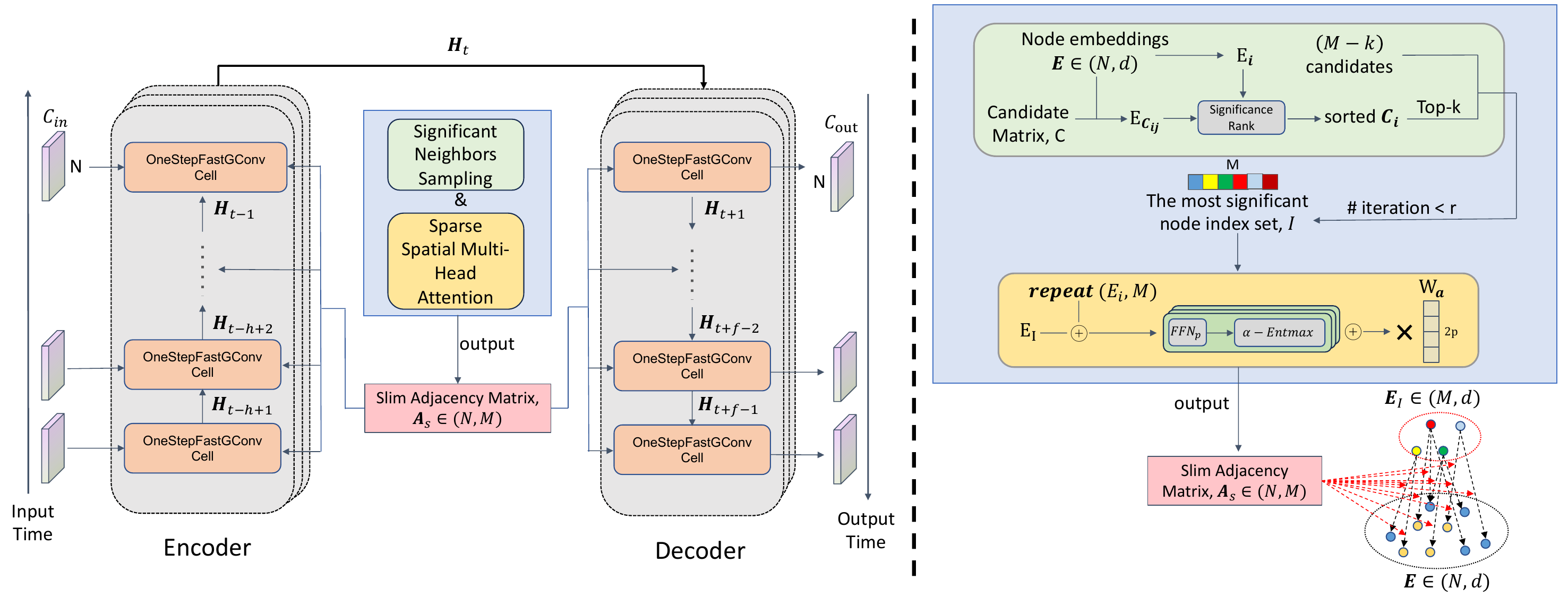}
  \caption{The overall architecture of the proposed SAGDFN framework. The Significant Neighbors Sampling module (green) and the Sparse Spatial Multi-Head Attention module (yellow) learn the slim adjacency matrix (pink) which will be employed in each OneStepFastGConv cell (orange).}
  \label{fig:SSTGDN}
\end{figure*}

Our Scalable Adaptive Graph Diffusion Forecasting Network (SAGDFN) aims to make the forecasting framework suitable for larger graphs by reducing computational and memory complexity. Compared to other approaches, our SAGDFN model possesses the capability to handle larger graphs, integrating beneficial information to enhance the model's forecasting accuracy. This is accomplished by the selective sampling of the most relevant neighbors, facilitating the creation of a slim dense adjacency matrix in a data-driven manner. To our knowledge, this represents the first investigation employing the $\alpha-Entmax$ function to refine adaptive spatial correlations. This novel application contributes significantly to improved forecasting performance (as shown in Table \ref{tab:ablation}). As a result, our approach achieves both efficient and effective performance for multivariate time series forecasting. In this section, we introduce the SAGDFN framework, and the detailed architecture of our proposed framework is shown in Figure \ref{fig:SSTGDN}. Our proposed SAGDFN is based on the classical encoder-decoder framework \cite{sutskever2014sequence} and we do not specify any prior knowledge about spatial correlation. Consequently, our SAGDFN model exhibits broad applicability to a wide spectrum of multivariate time series forecasting challenges, affirming its generic utility in this domain. The SAGDFN contains three main modules:  a \textit{most Significant Neighbors Sampling module} to derive the node index set $I$ that contains the indices of the most informative neighbors, a \textit{Sparse Spatial Multi-Head Attention module} to compute the slim dense adjacency matrix, and an \textit{Encoder-Decoder-based Forecasting module} to account for complex spatial-temporal correlation. These three modules are trained jointly in an end-to-end approach.

To reduce computational complexity and memory cost, we first propose the most Significant Neighbors Sampling module to filter and select the most important node indices from the node embedding matrix, resulting in an index set $I$ that contains the nodes index to facilitate the message passing function of Graph Neural Networks (GNNs) in the SAGDFN framework. The learned index set $I$ is then sent to the Sparse Spatial Multi-Head Attention module as input, which retrieves the embedding of the most significant neighbors $\mathbf{E}_I$ based on the index set $I$. The module then concatenates the node embedding matrix $\mathbf{E}$ with the most significant neighbors embedding $\mathbf{E}_I$ and employs a multi-head attention mechanism to compute the slim dense adjacency matrix $\mathbf{A}_s$. The slim dense adjacency matrix $\mathbf{A}_s$ is used by a fast graph convolution in the Encoder-Decoder-based forecasting module to model spatial correlation and temporal dependency. SAGDFN updates the most significant node index set $I$, the slim dense adjacency matrix $\mathbf{A}_s$, and all the learnable parameters $\Theta$, including the node embedding matrix $\mathbf{E}$, jointly in an end-to-end approach using backpropagation based on the L1 loss.

SAGDFN addresses the high computational complexity and memory cost issue of the STGNNs while ensuring high prediction accuracy.
SAGDFN learns to sample the most pertinent neighbors and captures the pair-wise spatial correlations in an end-to-end approach. Our method achieves state-of-the-art performance on both commonly used and large multivariate time series forecasting datasets.
\subsection{Significant Neighbors Sampling}
The present graph neural networks-based forecasting models~\cite{AGCRN, MTGNN, EnhanceNet, GTS} learn a full adjacency matrix $\mathbf{A}$ with size $N \times N$ adaptively. These models integrate the learned adjacency matrix with the subsequent graph convolution operation to offer structured spatial correlations that are useful for forecasting. However, such a manner involves quadratic computational complexity and becomes prohibitively expensive as $N$ grows over 2000, which is quite common in practical applications. To address this, we propose a most Significant Neighbors Sampling module, which is motivated by the observation~\cite{GRAPHWaveNet, EnhanceNet, GRIN, DGCNN} that not all nodes are important to a given node and only a small subset of nodes is significant to it, which we refer to as its most significant neighbors. Our proposed module takes advantage of such an observation and dynamically selects these significant neighbors out of the $N$ nodes. After that, the selected neighbors are utilized to construct a slim adjacency matrix that achieves a balance between fast computation and accurate forecasting performance.
\begin{algorithm}[h]
 \caption{Significant Neighbors Sampling\label{alg:neighbor-sampling}}
 \footnotesize
 \LinesNumbered
 \KwIn{Node Embedding: $\mathbf{E}$, Candidate Neighbors: $\mathbf{C}$}
 \KwOut{Significant Neighbor Index: ${I}$}
 $\mathbf{d} \leftarrow \mathbf{0}_{M}$\;
 \For {$i \leftarrow $ 1 to $N$} {
    \For {$j \leftarrow $ 1 to $M$} {
        $d_{j} \leftarrow \| \mathbf{E}_i - \mathbf{E}_{C_{ij}} \|_2$\;
    }
    Sort $\mathbf{C}_i$ by $\mathbf{d}$\; 
 }
 count $\leftarrow$ \textrm{Frequency of node ids in} $\mathbf{C}_{1:N, 1:K}$\;
 $V_K \leftarrow$ \textrm{Node ids with top}-$K$ count\;
 $V_{M-K} \leftarrow$ \textrm{Random sampling} $M - K$ \textrm{nodes from} $V \setminus V_K$\;
 $I \leftarrow V_K \cup V_{M-K}$\;
 \Return $I$\;
\end{algorithm}

Bearing this idea in mind, one may attempt to select the most significant neighbors that are specific to a particular node $i$ ($i = 1 \dots N$). Since the most significant neighbors vary across different nodes, this will result in a sparse adjacency matrix $\mathbf{A}$ which may not be as computationally efficient as the dense matrix for the present automatic differentiation platforms (PyTorch, TensorFlow, etc.). In contrast, we propose a more memory-efficient method to select the most significant $M \ll N$ neighbors that are shared by all nodes and then use these $M$ nodes to construct an adjacency matrix $\mathbf{A}_s$ with size $N \times M$ that will be used in the subsequent graph convolution and forecasting process. To this end, we randomly initialize a node embedding matrix $\mathbf{E} \in \mathbb{R}^{N \times d}$, whose $i$-th row is denoted by $\mathbf{E}_i \in \mathbb{R}^d$. We also randomly construct a candidate neighbors matrix $\mathbf{C} \in \{1,\ldots, N\}^{N \times M}$ whose $i$-th row indicates the most significant neighbor candidates of node $i$, with the requirement that each node id can only appear once in a row, ensuring that the amortized count of every distinct node is around $M$ in the candidate matrix, that is, every distinct node being considered by the candidate neighbors matrix. In other words, $\mathbf{C}_i$ is the neighbor candidate queue of node $i$. 

As shown in Algorithm~\ref{alg:neighbor-sampling}, given the present node embedding matrix $\mathbf{E}$, we first rank the significance of the $M$ candidate neighbors of node $i$ by measuring the similarity between node $i$ and the $j$-th candidate $C_{ij}$ in the representation space (lines 1-5 in Algorithm~\ref{alg:neighbor-sampling}), i.e., the closer the candidate is in the representation space, the more significant it is. Specifically, we adopt the Euclidean distance to measure the similarity between the time series pairs for its efficacy in the context of normalized representation spaces~\cite{Euclidean1, Euclidean2}. In this way, we have put the most significant neighbors of all nodes into the front of the candidate queues. To select the $M$ most significant neighbors that are globally influential and could be shared by all nodes, we measure the significance of each node by counting its frequency in the top-$K$ position of candidate neighbors matrix, i.e., $\mathbf{C}_{1:N, 1:K}$ ($K < M$), and only take the top-$K$ significant nodes, namely, $V_K$ (lines 6-7 in Algorithm~\ref{alg:neighbor-sampling}). Note that, in each Significant Neighbors Sampling iteration, we intentionally leave $M-K$ empty slots and fill them by randomly sampling from the set $V \setminus V_K$, in order to encourage exploration and enhance model robustness (line 8 in Algorithm~\ref{alg:neighbor-sampling}) so that eventually all node embeddings are updated and important variates are not inadvertently omitted. Eventually, the selected top-$M$ significant node index set $I$ is returned by the sampling algorithm. It will be used to construct a slim dense adjacency matrix $\mathbf{A}_s$ with size $N \times M$, which significantly reduces both the computational and memory complexity in comparison with the original $\mathcal{O}(N^2)$ implementation. 
\subsection{Sparse Spatial Multi-Head Attention}
Given the returned most significant neighbors index $I$ and node embedding matrix $\mathbf{E}$, we propose a Sparse Spatial Multi-Head Attention module to derive the required slim dense adjacency matrix $\mathbf{A}_s  \in \mathbb{R}^{N \times M}$ in this section. Our Sparse Spatial Multi-Head Attention module boasts two critical characteristics. First, it operates independently of any pre-existing spatial knowledge, ensuring that the model is uninfluenced by potentially misleading spatial information. This also renders our SAGDFN universally applicable to all facets of multivariate time series forecasting. Secondly, we have strategically engineered a sparse attention mechanism. This design employs the $\alpha-Entmax$ function to learn and refine spatial correlations, leading to a marked enhancement in forecasting performance. Intuitively, the entry $\mathbf{A}_{s-ij}$ quantifies the correlation strength between node $i$ and the $j$-th most significant neighbors, that is to say, how much information should diffuse from the $j$-th most significant neighbors to node $i$ in the graph convolution operation. To capture such correlation, it is natural to adopt the inner product operation, i.e., let $\mathbf{A}_{s-ij} \propto \mathbf{E}_i^\top \mathbf{E}_{I_j} $. The inner product permits matrix operation to exploit the computation parallelizability. However, it fails to reveal the nonlinear correlation, which has to be compensated by the subsequent nonlinear transformation and more considerable depth (as adopted in Transformer~\cite{Transformer}). Hence, we propose to directly learn the nonlinear correlation with the Feed Forward Network. For a given node $i$, its multi-head attention scores regarding the $M$ most significant neighbors are computed as follows:
\begin{alignat}{3}
\bar{\mathbf{E}}_i &= \oplus\left(\operatorname{repeat}(\mathbf{E}_i, M), \mathbf{E}_I\right)  &&\in \mathbb{R}^{M \times 2d} \\
\mathbf{Y}_i^p &= \operatorname{FFN}_p(\bar{\mathbf{E}}_i) &&\in \mathbb{R}^{M \times 2} \\
\mathbf{Z}_i^p &= \alpha\text{-}\operatorname{Entmax}(\mathbf{Y}_i^p)  &&\in \mathbb{R}^{M \times 2}\\
\mathbf{Z}_i &= \oplus(\mathbf{Z}_i^1, \mathbf{Z}_i^2, \ldots, \mathbf{Z}_i^P)  &&\in \mathbb{R}^{M \times 2P}
\end{alignat}

where $\oplus$ indicates the concatenation operation and $p$ ranges from $1$ to the number of heads $P$.

The $\operatorname{repeat}()$ operator first constructs a matrix with size $M \times d$ whose rows are all identical to $\mathbf{E}_i$, and we then concatenate it with the embeddings of the most significant neighbors $\mathbf{E}_I$ along the representation dimension to yield $\bar{\mathbf{E}}_i$. $\bar{\mathbf{E}}_i$ is then transformed nonlinearly by the Feed Forward Network ($\mathrm{FFN}$) to produce $\mathbf{Y}_i$, a two-column matrix, whose first (resp. second) column represents the likely (resp. unlikely) possibility of correlations between the node $i$ and significant neighbors. Inspired by the multi-head attention, we also yield multiple $\mathbf{Y}_i^p$ with $P$ Feed Forward Networks to explore diverse correlations in different semantic spaces. Next, rather than employing the commonly used Softmax function, we propose to leverage the $\alpha$-Entmax to normalize $\mathbf{Y}_i^p$ along the column dimension to encourage sparsity (the benefits will be discussed in detail later), and the normalized scores are denoted by $\mathbf{Z}_i^p$ which will be concatenated to yield the multiple-head score matrix $\mathbf{Z}_i \in \mathbb{R}^{M \times 2P}$ for the node $i$. 
In the end, we stack all multi-head attention scores $\mathbf{Z}_i$ to yield a tensor $\mathbf{Z} \in \mathbb{R}^{N \times M \times 2P}$, which will be transformed by a linear function parameterized by the matrix $\mathbf{W}_a \in \mathbb{R}^{2P \times 1}$ to produce the desired adjacency matrix $\mathbf{A}_s$ as follows.
\begin{alignat}{3}
\mathbf{Z} &= \operatorname{stack}([\mathbf{Z}_1, \mathbf{Z}_2, \ldots, \mathbf{Z}_N]) &&\in \mathbb{R}^{N \times M \times 2P} \\
\mathbf{A}_s &= \mathbf{Z} \mathbf{W}_a &&\in \mathbb{R}^{N \times M}
\end{alignat}

\noindent\textbf{Remark}. In the multivariate time series forecasting problem, the spatial correlation derived from the Softmax function \cite{AGCRN, MTGNN, EnhanceNet} comprise a significant proportion of low-weight entries. Consequently, directly performing the graph convolution based on these low-weight entries may distract the node of interest and weaken the impact of the graph convolution, as the nodes connected to the target node via such low-weight entries typically exhibit minimal or no trend and seasonality similarity. To tackle this concern, we propose to use the $\alpha$-Entmax~\cite{entmax2} with an adjustable hyperparameter $\alpha$ to suppress the impact from distant high-dimensional neighbors while accentuating the impact from proximate high-dimensional counterparts. Compared to the Softmax function which is typically employed to normalize the input vector and derive the approximation probability, the $\alpha$-Entmax function offers enhanced control over the normalized attention scores $\mathbf{Z}$ which serve as the spatial correlations. Its definition is given as follows, where $[x]_+ := max\{x, 0\}$:

\begin{equation}
\alpha\text{-Entmax}(\mathbf{z}) = [(\alpha-1)\mathbf{z}-\tau\mathbf{1}]^{1/\alpha-1}_+
\end{equation}\\
where $\mathbf{z}\in{\mathbb{R}^{d}}$ and $\tau:{\mathbb{R}^{d}}\rightarrow\mathbb{R}$ can be derived by the function below, for any $z$:
\begin{equation}
\sum_j \left[(\alpha-1)z_j - \tau(\mathbf{z})\right]^{\frac{1}{\alpha-1}}_+= 1
\end{equation}
The $\alpha$-Entmax function is a generalization of the Softmax with $\alpha = 1.0$ and the Sparsemax~\cite{Sparsemax} with $\alpha = 2.0$, and thus it offers more flexibility in comparison to these counterparts. The suppression and promotion level degree varies across different datasets, as different data domains contain their own inherent properties. The value of hyper-parameter $\alpha$, with a range of $[1.0, 2.5]$, can be tuned accordingly.

\subsection{Encoder-Decoder-based Forecasting}
Thus far, we have described how to adaptively derive the desired slim graph adjacency matrix $\mathbf{A}_s$ that not only provides rich spatial dependency relationships but also enables fast graph convolution operation with a small memory footprint. However, the dynamics of multivariate time series data is jointly controlled by the spatial correlation and temporal dependency, and thus it is equally important to model the temporal dynamics for accurate forecasting. To this end, we propose to integrate our fast graph convolution with a sequential model, more specifically, GRUs~\cite{GRU} to develop our entire forecasting model under the Encoder-Decoder architecture. First, we present how to perform fast graph convolution with our derived slim adjacency matrix $\mathbf{A}_s$ to diffuse valuable information across the spatial dimension to enhance the forecasting. Then, we describe the method of integrating this fast graph convolution operation with the GRU model to carry one-step computation, dubbed as \textrm{OneStepFastGConv} (One Step Fast Graph Convolution). In the end, we show how to make multi-step forecasting to produce the predictions and calculate the loss. The overall learning algorithm is shown in Algorithm~\ref{alg:overall-learning}, in which all the learnable parameters are denoted by set $\Theta$.
\begin{algorithm}[h]
 \caption{Overall Learning Algorithm\label{alg:overall-learning}}
 \footnotesize
 \LinesNumbered
 \KwIn{Training Dataset: $\mathcal{X}_h$, $\mathcal{X}_f$}
 \KwOut{Learned Parameters: ${\Theta}$}
 ${\Theta} \leftarrow$ \textrm{Randomly initializing all parameters}\;
 $\mathbf{C} \leftarrow$ \textrm{Randomly constructing candidates}\;
 
 $iter \leftarrow 0$\;
\While {training is true} {
     \If{$iter < r$}{$I \leftarrow \mathrm{SignificantNeighborSampling}(\mathbf{E}, \mathbf{C})$\;}
     
     $\mathbf{A}_s \leftarrow \mathrm{SparseSpatialMHAttention}(\mathbf{E}, I)$\;
     \tcp{Encoder: computing $\mathbf{H}_{t_0-1}$.}
     \For {$t \leftarrow t_0-h+1$ to $t_0-1$} {
        $\mathbf{H}_t, \hat{\mathbf{X}}_{t+1} \leftarrow \mathrm{OneStepFastGConv(\mathbf{A}_s, \mathbf{X}_t, \mathbf{H}_{t-1})}$\;
     }
     \tcp{Decoder: computing $\hat{\mathbf{X}}_{t_0+1}, \ldots, \hat{\mathbf{X}}_{t_0+f}$.}
     $\hat{\mathbf{X}}_{t_0} \leftarrow {\mathbf{X}}_{t_0}$\;
     \For {$t \leftarrow t_0$ to $t_0 + f$} {
        $\mathbf{H}_t, \hat{\mathbf{X}}_{t+1} \leftarrow \mathrm{OneStepFastGConv(\mathbf{A}_s, \hat{\mathbf{X}}_t, \mathbf{H}_{t-1})}$\;
     }
     {L} $\leftarrow$ \textrm{Compute loss with Equation}-\ref{eq:loss}\;
     Running backward computation to calculate $\partial L / \partial {\Theta}$\;
     Updating $\boldsymbol{\Theta}$ with $\partial L / \partial {\Theta}$\;
      $iter \leftarrow iter + 1$\;
 }
 \Return $\boldsymbol{\Theta}$\;
\end{algorithm}

We further conduct the information diffusion over the graph with multiple steps with adjacency matrix $\mathbf{A_s}$, and the corresponding fast graph convolution is defined as follows:
\begin{equation}
\begin{aligned}
 \mathbf{W}\star_{\mathbf{A}_s} \mathbf{X} = \sum_{j=0}^{J-1} \mathbf{W}_j(\mathbf{D}+\mathbf{I}_{N \times N})^{-1}(\mathbf{A}_s \mathbf{X}_{I}+\mathbf{X})^j
\end{aligned}
\end{equation}
where $\mathbf{W} = \{\mathbf{W}_j\}_j^J$ are the learnable parameters and $\mathbf{D} \in \mathbb{R}^{N\times N}$ is the degree matrix derived from $\mathbf{A}_s$, $J$ is the number of steps of graph diffusion. $\mathbf{X}$ denotes the observation at one particular time step, and we omit the subscript $t$ to keep the notation uncluttered. Note that the boldface symbol $\mathbf{I}_{N\times N}$ represents the N-dimension identity matrix whereas $I$ denotes the index set of the $M$ most significant neighbors. Hence, $\mathbf{A}_s \mathbf{X}_I$ represents the incoming information aggregated from the $M$ most significant neighbors, we add $\mathbf{A}_s \mathbf{X}_I$ with $\mathbf{X}$ to preserve the information of each node itself, and the sum is normalized by $(\mathbf{D}+\mathbf{I}_{N \times N})^{-1}$ before the linear transformation.

Our proposed approach is characterized by its generality and compatibility with a diverse set of models, including Recurrent Neural Networks (RNNs), Temporal Convolutional Networks (TCNs), and attention mechanisms. In particular, we seamlessly integrate our method into the classical Gated Recurrent Unit (GRU) architecture by strategically replacing the standard matrix multiplication with the graph convolution operation as follows:
\begin{equation}
\label{eq:one-step-fast-gconv}
\begin{aligned}
\mathbf{R}_t &= \sigma(\mathbf{W}_r\star_{\mathbf{A}_s}\oplus(\mathbf{X}_t, \mathbf{H}_{t-1})+\mathbf{b}_r) \\
\mathbf{Z}_t &= \sigma(\mathbf{W}_z\star_{\mathbf{A}_s}\oplus(\mathbf{X}_t, \mathbf{H}_{t-1})+\mathbf{b}_z) \\
\mathbf{\tilde{H}}_t &= \tanh(\mathbf{W}_{h}\mathbf{\star}_{\mathbf{A}_s}\oplus(\mathbf{X}_t, \mathbf{R}_t \odot \mathbf{H}_{t-1}) + \mathbf{b}_{h})) \\
\mathbf{H}_t &= \mathbf{Z}_t \odot \mathbf{H}_{t-1} + (1 - \mathbf{Z}_t) \odot \mathbf{\tilde{H}}_t \\
\hat{\mathbf{X}}_t &= \mathbf{H}_t \mathbf{W}_x
\end{aligned}
\end{equation}
where $\mathbf{X}_t$ and $\mathbf{H}_t$ represent the input and hidden state at time step $t$, $\mathbf{R}_t$ and $\mathbf{Z}_t$ denote the reset gate and update gate at time step $t$, respectively. $\sigma$ signifies the sigmoid activation function and $\mathbf{W}_r$, $\mathbf{W}_z$, $\mathbf{W}_{h}$ are learnable model parameters. The prediction $\mathbf{\hat{X}}_t$ at step $t$ is obtained by a linear transformation with parameter matrix $\mathbf{W}_x$. We refer to the computation of Equation~\ref{eq:one-step-fast-gconv} as \textrm{OneStepFastGConv}. To make the multiple-step predictions, we run \textrm{OneStepFastGConv} $h - 1$ steps with the historical observation sequence $\mathcal{X}_h$ as input to compute its compressed representation $\mathbf{H}_{t_0-1}$, i.e., the Encoder as shown in lines 8-9 in Algorithm~\ref{alg:overall-learning}; analogously, we run it $f$ steps to produce the prediction sequence $\hat{\mathbf{X}}_{t_0+1}, \ldots, \hat{\mathbf{X}}_{t_0+f}$ by using the obtained sequence representation $\mathbf{H}_{t_0-1}$ and the observation at $t_0$ step $\mathbf{X}_{t_0}$. 
Eventually, we calculate the prediction loss by adopting the commonly used MAE (Mean Absolute Error) as the metric, i.e.,
\begin{equation}
\label{eq:loss}
 \mathcal{L}(\mathbf{X}, \mathbf{\hat{X}}; \mathbf{\Theta}) = \frac{1}{N f} \sum_{t=t_{0}}^{t_{0}+f} \left\|\mathbf{X}_{t} - \mathbf{\hat{X}}_{t}\right\|_1
\end{equation}
Lines 13-15 in Algorithm~\ref{alg:overall-learning} show the optimization steps by calculating the gradients of all parameters with respect to the loss $L$. Note that the most significant neighbors and adjacency matrix are derived in lines 5-7
 in Algorithm~\ref{alg:overall-learning} with the given embedding matrix $\mathbf{E}$. As $\mathbf{E} \in \Theta$ its values will be updated in each iteration, and thus the index set $I$ and adjacency matrix $\mathbf{A}_s$ will also be optimized dynamically so as to better aid the forecasting. Once the updating process of the embedding matrix $\mathbf{E}$ converges, to produce better forecasting performance, we complete the significant neighbors sampling process and stop introducing randomly sampled nodes to the index set $I$. We use the hyper-parameter convergence iteration number $r$ to control this process.   

\section{Experiments}\label{sec:experiments}
In this section, we 
evaluate the effectiveness of our proposed SAGDFN method on four real-world datasets. We first introduce experimental settings and compare the performance of SAGDFN with baselines. Furthermore, we design experiments to verify the superiority of the Significant Neighbors Sampling module and the Sparse Spatial Multi-Head Attention module. These two key components effectively model the spatial correlations in a slim way while reducing computational complexity. Finally, we conduct ablation studies to evaluate the impact of the essential architectures, components, and training strategies. The source code is available at \url{https://github.com/JIANGYUE61610306/SAGDFN.git}.
\vspace{-0.5cm}
\begin{table}[hbt!] 
    \centering 
    \caption{Statistics of our datasets}\label{tab:Datasets}
    \begin{tabular}{cccc}  
        \hline 
        Data type&Datasets&\# of sensors&Time range\\
        \hline 
         \multirow{3}*{\makecell{Traffic \\speed}} & METR--LA & 207 & 1 Mar - 30 June 2012\\
          & London2000 & 2000 & 1 Jan - 31 Mar 2020\\
          & NewYork2000 & 2000 & 1 Jan - 31 Mar 2020\\
        \hline 
        \makecell{Carpark \\lots} & CARPARK1918 & 1918 & 1 May - 30 June 2021\\
        \hline 
    \end{tabular}
\end{table}
\subsection{Experimental Setup}
\noindent\textbf{Datasets}. We perform experiments on four real-world multivariate time series datasets, and the statistics of the datasets is presented in Table \ref{tab:Datasets}.  METR-LA is a commonly used multivariate time series benchmark dataset that comprises traffic speed recordings from 207 nodes with a time step of 5 minutes. To validate the strength of SAGDFN in tackling 
large
datasets, we introduce three real-world datasets: CARPARK1918, London2000, and NewYork2000. 
London2000 and NewYork2000 contain traffic speed hourly recordings of 2000 road segments, while CARPARK1918 contains parking lot availability information of 1918 carparks in Singapore with a time step of 5 minutes.
Note that previous work on multivariate time series prediction has not reported evaluation on datasets of such scales. We believe the three datasets provide a different perspective for evaluating new approaches on the topic. 


To ensure a fair comparison, we split the dataset following the setting applied in previous studies. Specifically, we use 70\% of the available data as the training set, 10\% as the validation set, and 20\% as the testing set. For the traffic speed datasets, namely METR-LA, London2000, and NewYork2000, we utilize the features from the previous 12 time steps as input to predict the traffic speed of the next 12 time steps. As for the carpark availability dataset CARPARK1918, we use the parking records of the previous 24 time steps (i.e., 2 hours) to predict the number of available parking lots in the next 12 time steps (i.e., 1 hour).

\noindent\textbf{Baselines.} We perform evaluation on multivariate time series forecasting problem with $\textbf{15 baselines}$. We note that the baselines we select are all publicly available, which enables us to report the performance on our three large-scale datasets based on their original implementations. Specifically, the baselines include both classical time series forecasting methods (ARIMA~\cite{ARIMA}, VAR~\cite{VAR}, SVR~\cite{SVR} and LSTM~\cite{LSTM}) and various state-of-the-art STGNN methods (Employing predefined spatial correlation: DCRNN~\cite{DCRNN}, STGCN~\cite{STGCN}, STSGCN~\cite{STSGCN}; Employing adaptive spatial correlation by inner product method: Graph WaveNet~\cite{GRAPHWaveNet}, AGCRN~\cite{AGCRN}, MTGNN~\cite{MTGNN}, D2STGNN~\cite{D2STGNN}; Employing adaptive spatial correlation by attention scores: GMAN~\cite{GMAN} ASTGCN~\cite{ASTGCN}; Employing adaptive spatial correlation by pair-wise information analysis: GTS~\cite{GTS}, STEP~\cite{STEP}).

\noindent\textbf{Metrics.}  To evaluate the performance of different models, we adopt three widely-used metrics for multivariate time series forecasting, including Mean Absolute Error (MAE), Root Mean Squared Error (RMSE), and Mean Absolute Percentage Error (MAPE). A lower score indicates a better performance. \\
\begin{figure}
  \centering
  \includegraphics[width=\linewidth]{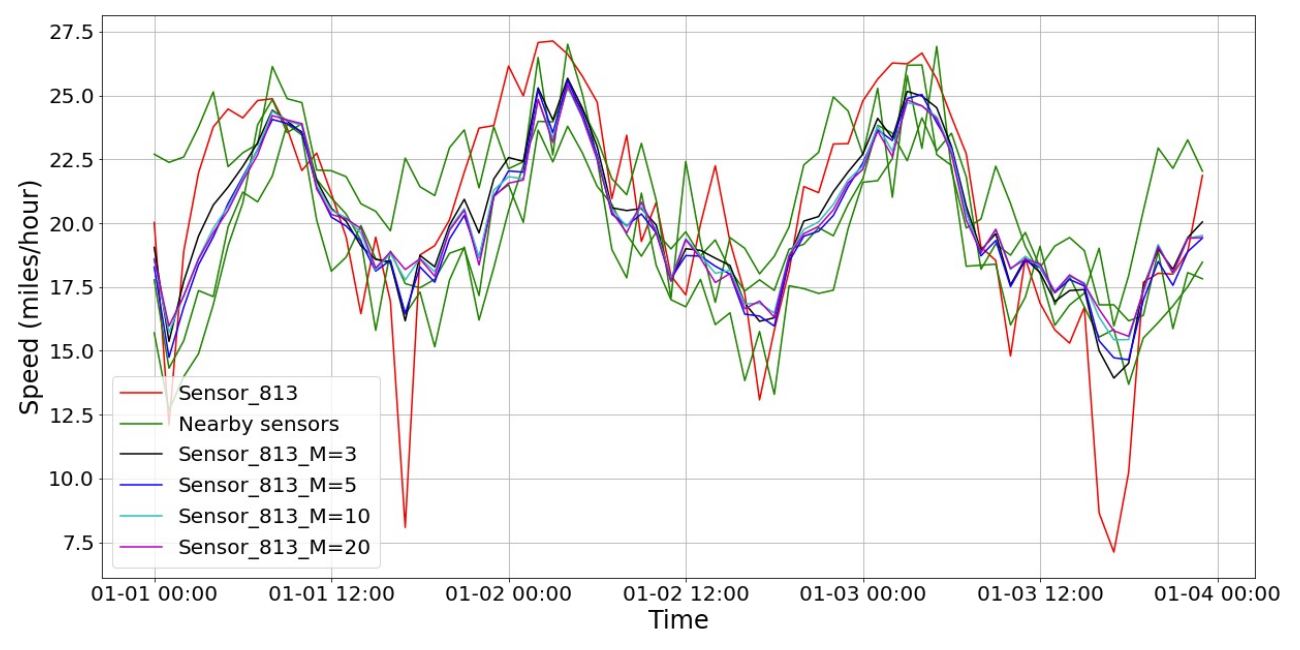}
  \caption{Diffusion threshold M for Sensor 883 of London2000 dataset.}
  \label{fig:diffusion_sizes}
\end{figure}
\begin{table*}[hbt!] 
    \centering 
    \caption{Performance comparison on METR--LA dataset.}\label{tab:METRLA}
    \vspace{-0.35cm}
    \begin{tabular}{c|ccc|ccc|ccccc}  
        \hline 
          \multirow{2}*{{METR--LA}}&\multicolumn{3}{c}{Horizon 3}&\multicolumn{3}{c}{Horizon 6}&\multicolumn{3}{c}{Horizon 12}\\ 
        &MAE& RMSE & MAPE &MAE& RMSE & MAPE&MAE& RMSE & MAPE\\ 
        \hline 
         ARIMA& 3.99 & 8.21& 9.6\% & 5.15 & 10.45 & 12.7\% & 6.9 & 13.23&17.4\%\\
         VAR & 4.42 &7.89&10.2\%&5.41&9.13&12.7\%&6.52	&10.11&15.8\%\\
         SVR& 3.99&8.45&9.3\%&5.05&10.87&12.1\%&6.72&13.76&16.7\%\\
         LSTM& 3.44&6.30&9.6\%&3.77&7.23&10.9\%&4.37&8.69&13.2\%\\
         DCRNN& 2.77&5.38&7.3\%&3.15&6.45&8.8\%&3.6&7.60&10.5\%\\
         STGCN& 2.88&5.74&7.6\%&3.47&7.24&9.6\%&4.59&9.4&12.7\%\\
         GRAPH WaveNet& 2.69&5.15&6.9\%&3.07&6.22&8.4\%&3.53&7.37&10.0\%\\
         GMAN& 2.80&5.55&7.4\%&3.12&6.49&8.7\%&3.44&7.35&10.0\%\\
         AGCRN& 2.87 & 5.58 & 7.7\% & 3.23 & 6.58 & 9.0\% & 3.62 & 7.51 & 10.4\%\\
         MTGNN& 2.69 & 5.18 & 6.9\% & 3.05 & 6.17 & 8.2\% & 3.49 & 7.23 & 9.9\%\\
         ASTGCN& 4.86 & 9.27 & 9.2\% & 5.43 & 10.61 & 10.1\% & 6.51 & 12.52 & 11.6\%\\
         STSGCN& 3.31 & 7.62 & 8.1\% & 4.13 & 9.77 & 10.3\% & 5.06 & 11.66 & 12.9\%\\
         \hline 
         GTS & 2.67 & 5.27 & 7.21\% & 3.04 & 6.25 & 8.4\% & 3.46 & 7.31 & 10.0\%\\
         STEP & 2.61 & 4.98 & \underline{\textbf{6.5\%}} & 2.96 & \underline{\textbf{5.97}} & 8.0\% & \underline{\textbf{3.37}} & \underline{\textbf{6.99}} & 9.6\%\\
         D2STGNN(c)& 2.57& \underline{\textbf{4.93}} & \underline{\textbf{6.5\%}} & \underline{\textbf{2.94}} & \underline{\textbf{5.97}} & \underline{\textbf{7.9\%}} & 3.41 & 7.15 & 9.6\%\\
         SAGDFN& \underline{\textbf{2.56}} & 5.00 & \underline{\textbf{6.5\%}} & \underline{\textbf{2.94}} & 6.05 & \underline{\textbf{7.9}}\% & \underline{\textbf{3.37}} & 7.17 & \underline{\textbf{9.5\%}}\\
         \hline
    \end{tabular}
    \vspace{-0.5cm}
\end{table*}

\begin{table*}[!htbp] 
    \centering  
    \caption{Performance comparison on London200 dataset. (AGCRN, GTS and D2STGNN are trained on data of maximum processable graph size as listed)}\label{tab:London200}
    \vspace{-0.35cm}
     \begin{tabular}{c|c|ccc|ccc|ccccc}  
        \hline 
          \multirow{2}*{{ London200}}&\multirow{2}*{{\# nodes in training set}}&\multicolumn{3}{c}{Horizon 3}&\multicolumn{3}{c}{Horizon 6}&\multicolumn{3}{c}{Horizon 12}\\ 
        & & MAE& RMSE & MAPE &MAE& RMSE & MAPE&MAE& RMSE & MAPE\\ 
        \hline 
         AGCRN & 1750&6.27& 8.58 & 26.2\%& 6.25& 8.56&26.1\% & 6.27&8.59&26.1\%\\
         GTS & 1000&2.16 & 3.23& 11.8\% & 2.56 & 3.69& 13.9\%& 2.05 & 3.03& 10.7\%\\
         D2STGNN(c)& 200&1.81&2.47&9.4\%&2.17&2.93&11.0\%&2.49&3.26&12.1\%\\
         \hline
         \multirow{4}*{{ SAGDFN}}& 200&1.75&2.46&9.4\%&1.97&2.78&10.7\%&2.28&3.14&12.4\%\\
         & 1000&1.61&2.25&8.6\%&1.82&2.55&10.0\%&2.11&2.87&11.3\%\\
         & 1750&1.53&2.15&8.3\%&1.71&2.39&9.2\%&1.97&2.72&10.7\%\\
          &5000& \underline{\textbf{1.44}} & \underline{\textbf{2.03}} &\underline{\textbf{7.8\%}} &\underline{\textbf{1.64}} &\underline{\textbf{2.29}} &\underline{\textbf{8.8\%}} &\underline{\textbf{1.89}} &\underline{\textbf{2.60}}&\underline{\textbf{10.2\%}}\\
         \hline
    \end{tabular}
    \vspace{-0.5cm}
\end{table*}
\vspace{-0.2cm}
\noindent\textbf{Implementation.} We evaluate all the models on a Linux workstation with an Intel(R) Xeon(R) Gold 6248 CPU @ 2.50GHz and a 32 GB Tesla V100 GPU. For all the baselines, we use their original implementations with minimum modifications required to run on our datasets. For our SAGDFN model, we use Adam optimizer and the nodes dimension $d$ is set to 100. As demonstrated in Figure \ref{fig:diffusion_sizes}, features of Sensor 813 merely change when increasing the number of nearby nodes $M$ from 10 to 20. In an attempt to ensure a wider margin, we set the latent neighborhood size, $M$, to 100, based on empirical evidence that performance stabilizes when increasing the size of $M$ to approximately $5\%$ of $N$, which rounds to 100. From all available candidates, we select the top-K = 80 nodes and randomly sample an additional $(M-K)=20$ nodes. The encoder-decoder forecasting layer is set to 1 and the depth of the Recurrent Graph Convolution Cell J is set to 3. The hidden size of GRUs is set to 64 and the default number of attention heads is 8.
We discuss the number of attention heads and the $\alpha$ value of the Entmax function in the Parameter Sensitivity section. 
To replicate the performance of the baseline models, we adopt the same hyper-parameter configurations recommended in their original papers. We report the performance of all the models using either a batch size of 64 or 32, depending on which one yields better results. In the case that a model still encounters out-of-memory (OOM) issue under a batch size of 32, we tag such models with an OOM symbol in the performance Table \ref{tab:CARPARK} to Table \ref{tab:NewYork}.
\begin{table*}[h] 
    \centering  
    \caption{\centering{Performance comparison on CARPARK1918 dataset. (Results marked $'\times'$ for model encountering the OOM issue)}}\label{tab:CARPARK}
    \vspace{-0.35cm}
    \begin{tabular}{c|ccc|ccc|ccccc}  
        \hline 
          \multirow{2}*{{ CARPARK1918}}&\multicolumn{3}{c}{Horizon 3}&\multicolumn{3}{c}{Horizon 6}&\multicolumn{3}{c}{Horizon 12}\\ 
        &MAE& RMSE & MAPE &MAE& RMSE & MAPE&MAE& RMSE & MAPE\\ 
        \hline 
         ARIMA& 3.31&7.41&7.4\%&5.59&10.03&12.1\%&9.15&16.68&14.1\%\\
         VAR& 5.45&10.81&13.5\%&7.41&14.63&19.9\%&10.65&20.17&27.3\%\\
         SVR& 17.71&47.42&16.11\%&19.35&47.93&20.2\%&21.92&49.00&22.3\%\\
         LSTM& 3.15&8.90&7.0\%&5.48&12.29&11.7\%&9.23&18.92&16.0\%\\
         DCRNN & 2.59&7.69&7.0\%&4.18&10.09&10.9\%&6.31&14.19&14.0\%\\
         STGCN & $\times$ & $\times$ & $\times$ & $\times$  & $\times$ & $\times$ & $\times$  & $\times$ & $\times$\\
         GRAPH WaveNet& 7.84 & 17.26&15.9\%&8.67&19.04&16.8\%&10.49&22.75&19.8\%\\
         GMAN & $\times$ & $\times$ & $\times$ & $\times$  & $\times$ & $\times$ & $\times$  & $\times$ & $\times$\\
         AGCRN & $\times$ & $\times$ & $\times$ & $\times$  & $\times$ & $\times$ & $\times$  & $\times$ & $\times$\\
         MTGNN&3.74&9.88&8.6\%&4.97&12.33&12.2\%&7.57&17.02&15.8\%\\
         ASTGCN & $\times$ & $\times$ & $\times$ & $\times$  & $\times$ & $\times$ & $\times$  & $\times$ & $\times$\\
         STSGCN & $\times$ & $\times$ & $\times$ & $\times$  & $\times$ & $\times$ & $\times$  & $\times$ & $\times$\\
         \hline 
         GTS & $\times$ & $\times$ & $\times$ & $\times$  & $\times$ & $\times$ & $\times$  & $\times$ & $\times$\\
         STEP & $\times$ & $\times$ & $\times$ & $\times$  & $\times$ & $\times$ & $\times$  & $\times$ & $\times$\\
         D2STGNN(c) & $\times$ & $\times$ & $\times$ & $\times$  & $\times$ & $\times$ & $\times$  & $\times$ & $\times$\\
         SAGDFN & \underline{\textbf{2.26}} & \underline{\textbf{7.74}} &\underline{\textbf{6.0\%}} &\underline{\textbf{3.77}} &\underline{\textbf{10.07}} &\underline{\textbf{10.2\%}} &\underline{\textbf{5.36}} &\underline{\textbf{13.17}}&\underline{\textbf{12.7\%}}\\
         \hline
    \end{tabular}
    \vspace{-0.5cm}
\end{table*}

\begin{table*}[h] 
    \centering  
    \caption{\centering{Performance comparison on London2000 dataset. (Results marked $'\times'$ for model encountering the OOM issue)}}\label{tab:London}
    \vspace{-0.35cm}
    \begin{tabular}{c|ccc|ccc|ccccc}  
        \hline 
          \multirow{2}*{{ London2000}}&\multicolumn{3}{c}{Horizon 3}&\multicolumn{3}{c}{Horizon 6}&\multicolumn{3}{c}{Horizon 12}\\ 
        &MAE& RMSE & MAPE &MAE& RMSE & MAPE&MAE& RMSE & MAPE\\ 
        \hline 
         ARIMA& 2.04&2.77&9.9\%&2.30&3.14&11.1\%&2.20&3.00&10.9\%\\
         VAR& 2.09&2.84&10.8\%&2.28&3.07&11.5\%&2.56&3.43&12.3\%\\
         SVR&2.66&3.72&13.1\%&3.16&4.42&15.2\%&3.35&4.75&16.0\%\\
         LSTM & 1.92&2.70&10.5\%&2.21&3.06&12.1\%&1.96&2.77&10.8\%\\
         DCRNN & 1.88&2.65&10.1\%&2.06&2.89&11.0\%&1.98&2.78&10.5\%\\
         STGCN & $\times$ & $\times$ & $\times$ & $\times$  & $\times$ & $\times$ & $\times$  & $\times$ & $\times$\\
         GRAPH WaveNet& 1.92&2.72&10.7\%&2.17&3.05&12.1\%&1.92&2.74&10.6\%\\
         GMAN & $\times$ & $\times$ & $\times$ & $\times$  & $\times$ & $\times$ & $\times$  & $\times$ & $\times$\\
         AGCRN & $\times$ & $\times$ & $\times$ & $\times$  & $\times$ & $\times$ & $\times$  & $\times$ & $\times$\\
         MTGNN&5.40&7.31&26.6\%&5.39&7.31&26.4\%&5.39&7.31&26.2\%\\
         ASTGCN & $\times$ & $\times$ & $\times$ & $\times$  & $\times$ & $\times$ & $\times$  & $\times$ & $\times$\\
         STSGCN & $\times$ & $\times$ & $\times$ & $\times$  & $\times$ & $\times$ & $\times$  & $\times$ & $\times$\\
         \hline 
         GTS & $\times$ & $\times$ & $\times$ & $\times$  & $\times$ & $\times$ & $\times$  & $\times$ & $\times$\\
         STEP & $\times$ & $\times$ & $\times$ & $\times$  & $\times$ & $\times$ & $\times$  & $\times$ & $\times$\\
         D2STGNN(c) & $\times$ & $\times$ & $\times$ & $\times$  & $\times$ & $\times$ & $\times$  & $\times$ & $\times$\\
         SAGDFN & \underline{\textbf{1.44}} & \underline{\textbf{2.05}} &\underline{\textbf{7.7\%}} &\underline{\textbf{1.63}} &\underline{\textbf{2.32}} &\underline{\textbf{8.7\%}} &\underline{\textbf{1.87}} &\underline{\textbf{2.66}}&\underline{\textbf{10.0\%}}\\
         \hline
    \end{tabular}
    \vspace{-0.5cm}
\end{table*}
\subsection{Performance of SAGDFN}
Table \ref{tab:METRLA} shows the performance of all the models at different forecasting horizons on METR-LA dataset. The best result for each evaluation metric is highlighted in bold with an underline. We observe that the proposed SAGDFN exhibits comparable performance with the state-of-the-art baselines. Specifically, SAGDFN achieves the best performance on 6 out of 9 evaluation metrics on the METR-LA dataset, which is higher than STEP \cite{STEP} and D2STGNN(c) \cite{D2STGNN}. Since the original implementation of D2STGNN\cite{D2STGNN} integrates additional day-in-week information in the estimation gate, we modify the implementation to align it with other methods by removing this factor and report the performance as D2STGNN(c).
Based on the results obtained from the METR-LA dataset, we observe that: 1) Deep learning methods generally demonstrate superior performance in comparison to non-deep learning methods, and incorporating the graph structure into the deep learning models can further enhance their performance. 2) Models that learn the adjacency matrix demonstrate superior performance over models that directly employ the predefined adjacency matrix, irrespective of the approach used to utilize the learned adjacency matrix. SAGDFN is solely based on the learned adjacency matrix, whereas models like GTS and STEP utilize the learned adjacency as a regularization term in the loss function, and D2STGNN employs both the learned adjacency matrix and the predefined adjacency matrix in GNNs. All of these models outperform DCRNN, which directly employs the predefined adjacency matrix. 3) Models that exploit the pairwise information among multivariate time series to derive spatial correlation typically outperform models that model the spatial correlation by taking the inner product of the nodes embeddings. For instance, GTS, STEP, and SAGDFN exhibit superior performance over models such as AGCRN and MTGNN.

\begin{table*}[h] 
    \centering  
    \caption{\centering{ Performance comparison on NewYork2000 dataset. (Results marked $'\times'$ for model encountering the OOM issue)}}\label{tab:NewYork}
    \vspace{-0.35cm}
    \begin{tabular}{c|ccc|ccc|ccccc}  
        \hline 
          \multirow{2}*{{ NewYork2000}}&\multicolumn{3}{c}{Horizon 3}&\multicolumn{3}{c}{Horizon 6}&\multicolumn{3}{c}{Horizon 12}\\ 
        &MAE& RMSE & MAPE &MAE& RMSE & MAPE&MAE& RMSE & MAPE\\ 
        \hline 
         ARIMA & 2.36&3.32&11.0\%&2.70&3.75&12.7\%&2.60&3.69&12.1\%\\
         VAR& 2.24&3.12&10.4\%&2.52&3.53&11.3\%&2.75&3.80&12.5\%\\
         SVR& 3.46&4.85&14.7\%&4.18&5.88&17.2\%&4.26&6.05&17.4\%\\
         LSTM & 2.49&3.50&12.2\%&2.87&3.93&13.8\%&2.37&3.36&11.5\%\\
         DCRNN & 2.41&3.41&11.6\%&2.96&4.15&13.5\%&2.99&4.01&13.5\%\\
         STGCN & $\times$ & $\times$ & $\times$ & $\times$  & $\times$ & $\times$ & $\times$  & $\times$ & $\times$\\
         GRAPH WaveNet & 2.32&3.27&11.7\%&2.60&3.65&13.3\%&\underline{\textbf{2.36}}&\underline{\textbf{3.30}}&11.5\%\\
         GMAN & $\times$ & $\times$ & $\times$ & $\times$  & $\times$ & $\times$ & $\times$  & $\times$ & $\times$\\
         AGCRN & $\times$ & $\times$ & $\times$ & $\times$  & $\times$ & $\times$ & $\times$  & $\times$ & $\times$\\
         MTGNN&7.26&11.27&29.4\%&7.259&11.28&29.1\%&7.26&11.30&28.9\%\\
         ASTGCN & $\times$ & $\times$ & $\times$ & $\times$  & $\times$ & $\times$ & $\times$  & $\times$ & $\times$\\
         STSGCN & $\times$ & $\times$ & $\times$ & $\times$  & $\times$ & $\times$ & $\times$  & $\times$ & $\times$\\
         \hline 
         GTS & $\times$ & $\times$ & $\times$ & $\times$  & $\times$ & $\times$ & $\times$  & $\times$ & $\times$\\
         STEP & $\times$ & $\times$ & $\times$ & $\times$  & $\times$ & $\times$ & $\times$  & $\times$ & $\times$\\
         D2STGNN(c) & $\times$ & $\times$ & $\times$ & $\times$  & $\times$ & $\times$ & $\times$  & $\times$ & $\times$\\
         SAGDFN & \underline{\textbf{1.81}} & \underline{\textbf{2.52}} &\underline{\textbf{8.9\%}} &\underline{\textbf{2.04}} &\underline{\textbf{2.84}} &\underline{\textbf{9.9\%}} &2.41&3.39&\underline{\textbf{11.1\%}}\\
         \hline 
    \end{tabular}
    \vspace{-0.5cm}
\end{table*}
\begin{table*}[h] 
    \centering  
    \caption{Ablation Study on CARPARK1918 dataset.}\label{tab:ablation}
    \vspace{-0.35cm}
    \begin{tabular}{c|ccc|ccc|ccccc}  
        \hline 
          \multirow{2}*{{ CARPARK1918}}&\multicolumn{3}{c}{Horizon 3}&\multicolumn{3}{c}{Horizon 6}&\multicolumn{3}{c}{Horizon 12}\\ 
        &MAE& RMSE & MAPE &MAE& RMSE & MAPE&MAE& RMSE & MAPE\\ 
        \hline 
         SAGDFN & \underline{\textbf{2.26}} & \underline{\textbf{7.74}} &\underline{\textbf{6.0\%}} &\underline{\textbf{3.77}} &\underline{\textbf{10.07}} &\underline{\textbf{10.2\%}} &\underline{\textbf{5.36}} &\underline{\textbf{13.17}}&\underline{\textbf{12.7\%}}\\
         w/o Entmax & 2.80&8.47&6.6\%&4.37&11.14&11.1\%&6.92&16.12&14.8\%\\
         w/o Attention & 2.35&7.89&6.1\%&4.00&10.42&10.5\%&6.01&14.25&13.5\%\\
         w/o SNS  & 2.68 & 8.07 & 8.2\%& 4.23  & 10.11 & 12.0\% & 6.30  & 14.18 & 15.0\%\\
         w/o SNS \& SSMA & 2.76 & 8.53 & 6.5\% & 4.37  & 11.12 & 11.0\% & 6.98  & 16.10 & 14.9\%\\
         \hline 
    \end{tabular}
    \vspace{-0.5cm}
\end{table*}
The aforementioned deep learning methods have shown excellent performance on the commonly used METR-LA dataset. However, they encounter significant computational challenges and memory issues from the GNNs when handling larger datasets such as CARPARK1918, London2000, and NewYork2000. Specifically, we first demonstrate that training on large graphs could provide more information about the underlying spatial correlation to further improve the final forecasting performance. Under the same batch size of 64, AGCRN, GTS and D2STGNN can only run on 1750, 1000, and 200 nodes respectively, without out of memory issue. To compare the forecasting performance, we split a London200 dataset, the subset of the London2000 dataset containing only 200 sensors for evaluation. As shown in Table \ref{tab:London200}, our SAGDFN outperforms the SOTA methods by a large margin when incorporating a larger graph in the model training stage and the forecasting performance of SAGDFN on the specific London200 dataset increases with more time series considered in the training process. Next, we present the metrics of all the models at different forecasting horizons on the three large-scale datasets in Table \ref{tab:CARPARK}, Table \ref{tab:London} and Table \ref{tab:NewYork} respectively. The results show that SAGDFN consistently outperforms baselines by a significant margin, whereas the state-of-the-art models are not capable of performing forecasting tasks on three large-scale datasets due to the out-of-memory issue. For instance, GTS and STEP models, which rely on pairwise information among different entities, incur extra computational costs as compared to directly computing the inner product of the nodes embeddings, thus facing the bottleneck when extending to larger datasets. Although GRAPH WaveNet and MTGNN do not encounter the out-of-memory issue, they are unable to surpass other baselines due to the challenges associated with approximating the spatial correlation using the inner product of the node embeddings when dealing with large graph structures. To facilitate a comprehensive performance assessment of our SAGDFN model, we have expanded our experimental purview to include non-GNN-based MTS forecasting methodologies, specifically TimesNet~\cite{TimesNet}, FEDformer~\cite{Fedformer}, and ETSformer~\cite{ETSformer}. These evaluations were conducted on the METR-LA dataset and the CARPARK1918 dataset. Notably, while TimesNet, FEDformer, and ETSformer exhibit commendable efficacy in long sequence forecasting tasks, they inherently lack mechanisms to consider the intricate spatial correlations between the time sequences. This omission leads to their consistent underperformance in comparison to our SAGDFN method, as presented in Table \ref{tab:non-GNN}. In contrast, our proposed SAGDFN effectively captures complex spatial-temporal correlation and is scalable to large-scale datasets without a predefined graph adjacency matrix, offering promising forecasting performance across all datasets.
\begin{table*}[h] 
    \centering  
    \caption{\centering{Performance comparison with non-GNN-based methods}}\label{tab:non-GNN}
    \begin{tabular}{c|ccc|ccc|ccccc}  
        \hline 
          \multirow{2}*{{ METR-LA}}&\multicolumn{3}{c}{Horizon 3}&\multicolumn{3}{c}{Horizon 6}&\multicolumn{3}{c}{Horizon 12}\\ 
        &MAE& RMSE & MAPE &MAE& RMSE & MAPE&MAE& RMSE & MAPE\\ 
        \hline 
         TimesNet& 4.27 & 9.89 & 9.8\% & 5.40 & 12.34 & 12.0\% & 7.17 & 15.35 & 15.6\%\\
         FEDformer& 5.02 & 10.12 & 11.2\% & 6.04 & 12.27 & 13.0\% & 7.94 & 15.29 & 16.7\%\\
         ETSformer& 5.72 & 10.21 & 11.4\% & 6.75 & 13.1 & 13.0\% & 8.59 & 14.62 & 16.0\%\\
         SAGDFN & \underline{\textbf{2.56}} & \underline{\textbf{5.00}} & \underline{\textbf{6.5\%}} & \underline{\textbf{2.94}} & \underline{\textbf{6.05}} & \underline{\textbf{7.9}}\% & \underline{\textbf{3.37}} & \underline{\textbf{7.17}} & \underline{\textbf{9.5\%}}\\
         \hline 
          \multirow{2}*{{ CARPARK1918}}&\multicolumn{3}{c}{Horizon 3}&\multicolumn{3}{c}{Horizon 6}&\multicolumn{3}{c}{Horizon 12}\\ 
        &MAE& RMSE & MAPE &MAE& RMSE & MAPE&MAE& RMSE & MAPE\\ 
         \hline
         TimesNet& 4.55 & 11.47 & 11.3\% & 5.41 & 13.42 & 12.7\% & 7.36 & 17.53 & 15.5\%\\
         FEDformer& 6.52 & 13.05 & 17.7\% & 6.84 & 14.11 & 17.9\% & 9.02 & 18.53 & 21.2\%\\
         ETSformer& 9.55 & 17.20 & 29.2\% & 10.80 & 19.61 & 33.4\% & 12.99 & 23.55 & 38.5\%\\
         SAGDFN & \underline{\textbf{2.26}} & \underline{\textbf{7.74}} &\underline{\textbf{6.0\%}} &\underline{\textbf{3.77}} &\underline{\textbf{10.07}} &\underline{\textbf{10.2\%}} &\underline{\textbf{5.36}} &\underline{\textbf{13.17}}&\underline{\textbf{12.7\%}}\\
         \hline
    \end{tabular}
    \vspace{-0.6cm}
\end{table*}
\subsection{Computation Cost}
To assess computational expense, we contrast the parameter count and training duration of the SAGDFN model against several others: DCRNN, AGCRN, MTGNN, GTS, and D2STGNN. These comparisons are performed on the CARPARK1918 dataset, with details documented in Table \ref{tab:computationcost}. Due to GPU memory constraints, all models but the MTGNN operate with a reduced batch size. Due to high computational costs, all baseline models have slower training speeds, and barring DCRNN, require an elevated quantity of parameters to grasp spatial correlations. Both AGCRN and MTGNN models have the ability to generate predictions spanning multiple steps ahead in a single run, resulting in less time required for inference. Compared to baseline models, our SAGDFN model demonstrates significant performance enhancements (as shown in Tables \ref{tab:METRLA} to \ref{tab:NewYork}), while demanding the lowest computational costs for both training and inference.
\vspace{-0.1cm}
\begin{table}[h] 
    \centering  
    \caption{The computation cost on the CARPARK1918 dataset}\label{tab:computationcost}
    \begin{tabular}{c|c|c|c}  
        \hline 
        \multirow{2}*{{ Model}}&\multirow{2}*{{\# Parameters}}&Training Time&Inference Time\\ 
        & & (s/epoch) & (s)\\ 
         \hline 
         DCRNN & 372353 & 3522.53 & 236.42\\
         \hline 
         AGCRN & 15345380 & 1256.17 & 57.46\\
         \hline 
         MTGNN & 17947612 & 684.53 & 34.17\\
         \hline 
         GTS & 19771091 & 2327.62 & 123.37\\
         \hline 
         D2STGNN & 426182 & 2864.03 & 133.84\\
         \hline 
         SAGDFN & 142377 & 341.27 & 18.93\\
         \hline 
    \end{tabular}
\end{table}
\subsection{Ablation Study}
In this subsection, we conduct an ablation study on the CARPARK1918 dataset by removing different components in our proposed SAGDFN to demonstrate their contributions to the model performance. To facilitate a systematic evaluation, we name five variants of our SAGDFN model as follows:
\begin{itemize}
    \item
    \textbf{w/o Entmax}: we replace the $\alpha$-Entmax function in SAGDFN model with a standard Softmax function in the Sparse Spatial Multi-Head Attention module.
    \item
    \textbf{w/o Pair-Wise Attention}: we remove the pair-wise multi-heads attention mechanism in SAGDFN model. We derive the slim adjacency matrix, $\mathbf{A}_s$ using the inner product of the node embedding, $\mathbf{E}$ and the transpose of the neighborhood latent embeddings, denoted as $\mathbf{E}_I^T$.
    \item
    \textbf{w/o SNS}: we remove the Significant Neighbors Sampling module in SAGDFN model. Instead of sampling from the Significant Neighbors Sampling module, we randomly generate the significant node index set, $\mathbf{I}$.
    \item
    \textbf{w/o SNS \& SSMA}: we remove the Significant Neighbors Sampling module and the Sparse Spatial Multi-Head Attention module in the SAGDFN model. We model the adjacency matrix based on the topological structure similar to \cite{DCRNN}. We retain the top-100 closer neighbors for each node in the derived adjacency matrix and set other entries to zero.
\end{itemize}
 To ensure a rigorous and fair evaluation of the SAGDFN, we conduct the experiments for the model variants using the same settings as the original SAGDFN model.
 The results in Table \ref{tab:ablation} show that SAGDFN consistently outperforms all variants, which validates the effectiveness of our proposed components. We have several observations as follows: 1) employing the Significant Neighbors Sampling module and the Sparse Spatial Multi-Head Attention module is effective; 2) the $\alpha$-Entmax function plays a pivotal role in mitigating the influence of noise nodes by actively suppressing their information propagation process and thereby enhancing our model’s learning process; 3) the pair-wise spatial information modeling is a better approach compared to the inner product similarity, as SAGDFN outperforms SAGDFN w/o Pair-Wise Attention, which further bolster forecasting performance. Last, the special OneStepFastGConv function enables the information flow over the sampled graph in an end-to-end manner, ensuring robustness in our predictions.
 Overall, the results demonstrate the superiority of our proposed SAGDFN model design.

\subsection{Parameter Sensitivity}
\begin{figure}
  \centering
  \includegraphics[width=1.0\linewidth]{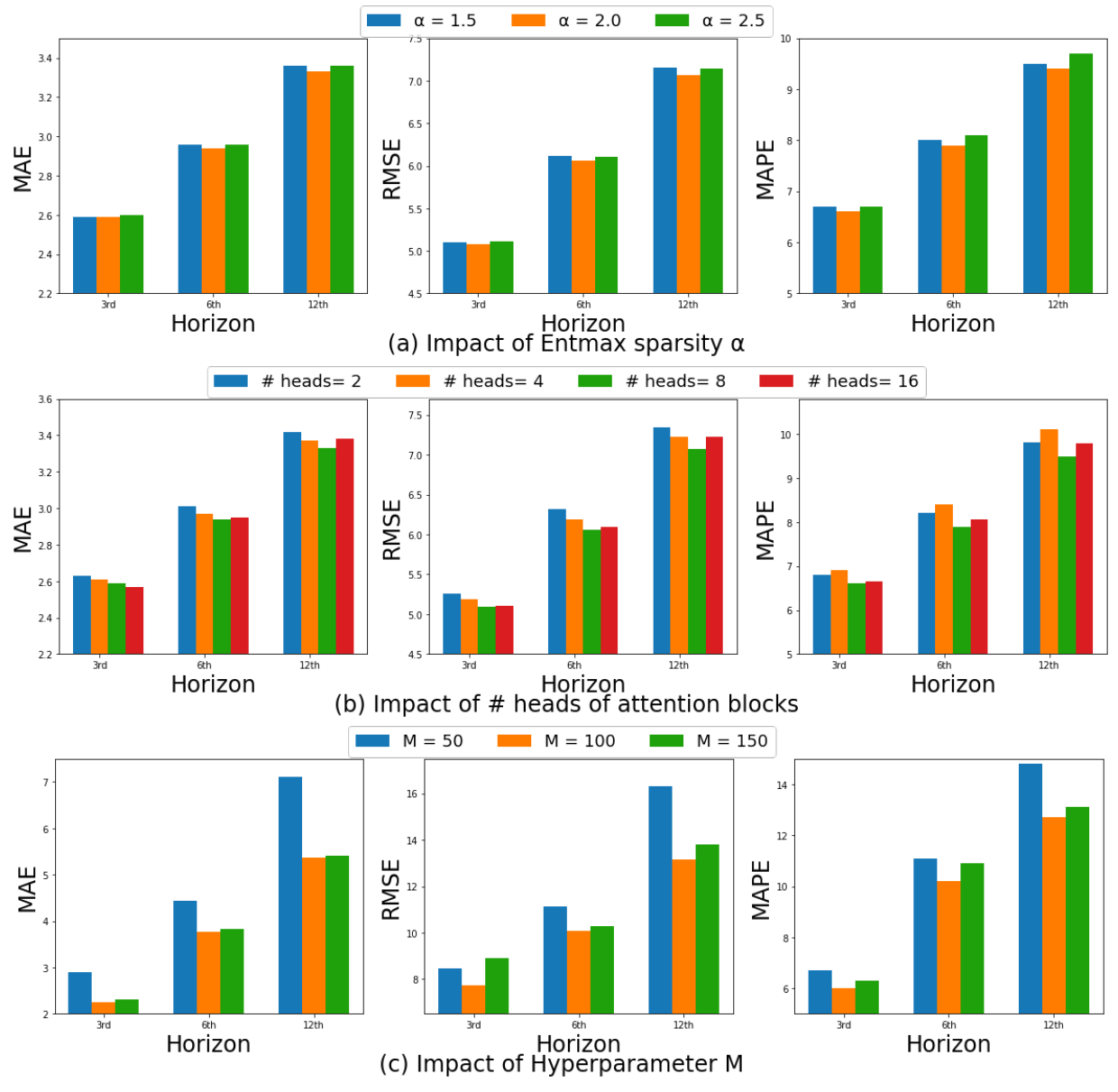}
  \caption{Hyper-parameter study.}
  \label{fig:hyperparameter}
\end{figure}

We further investigate the effects of the aforementioned hyper-parameters, namely the $\alpha$ value of the $\alpha$-Entmax function, the number of heads in the Sparse Spatial Multi-Head Attention module, and the size of the significant index set, $M$. We conduct experiments by varying these parameters while keeping all other factors untouched. The experimental results on the METR-LA dataset and the CARPARK1918 dataset are depicted in Figure \ref{fig:hyperparameter}. From the METR-LA dataset result, we can observe that the performance of the model improves as the number of heads used increases, albeit at the expense of higher computational costs. The model equipped with eight attention heads achieves the best performance. Moreover, our analysis reveals that a greater $\alpha$ value effectively filters out noise features with low spatial correlation, thus subsequently enhancing the model performance. Nonetheless, employing an excessively elevated $\alpha$ value risks missing nodes that contain valuable information. This would consequently lead to a performance drop in the model. As demonstrated in Figure \ref{fig:hyperparameter}(a), the optimal $\alpha$ value is determined to be 2.0. To study the impact of the hyper-parameter $M$, we conduct experiments on the CARPARK1918 dataset. As shown in Figure \ref{fig:hyperparameter}(c), increasing the size of the significant neighbors improves the model's performance in the early stages. However, performance becomes stable once a sufficient number of significant neighbors are taken. Further increasing the significant neighbor size merely wastes memory consumption without any performance improvement.
\subsection{Efficiency \& Visualization}
\begin{figure}
  \centering
  \includegraphics[scale=0.4]{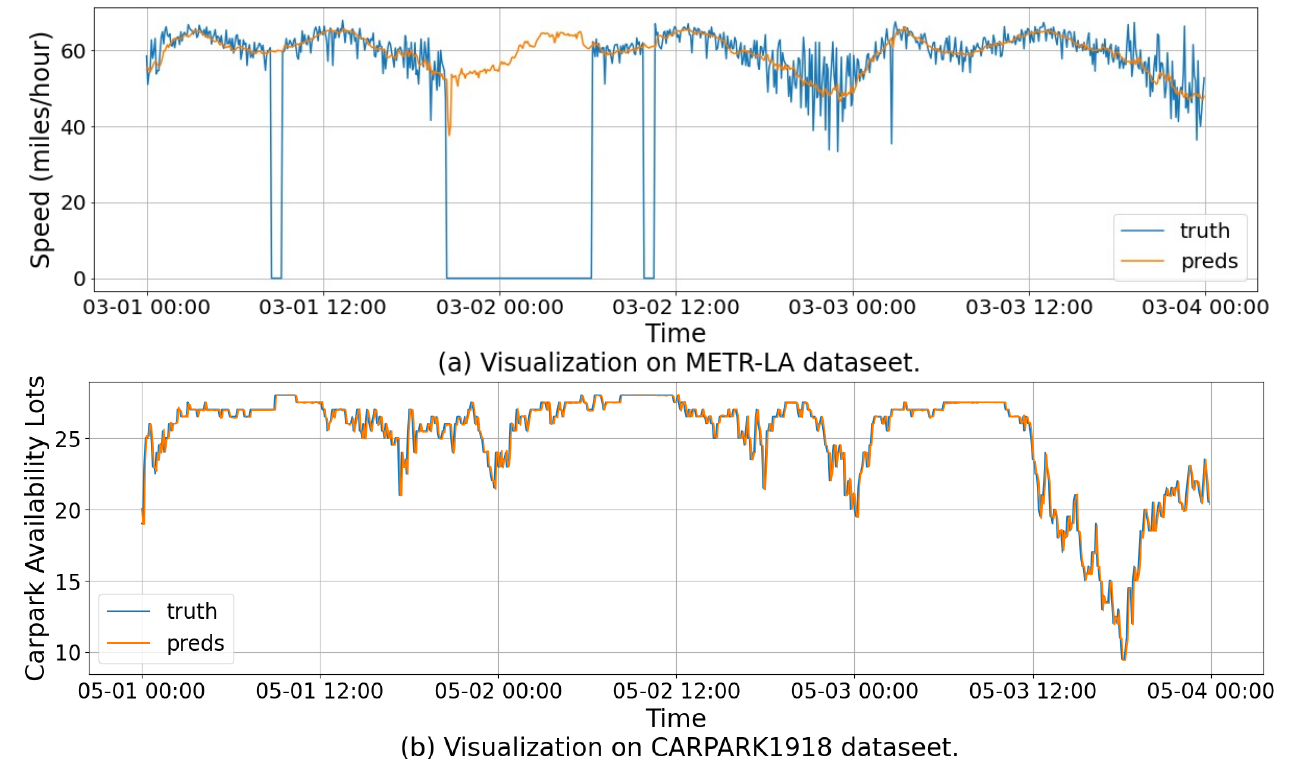}
  \vspace{-0.3cm}
  \caption{Visualizations on the METR-LA \& CARPARK1918 datasets.}
  \label{fig:visual}
\end{figure}

To gain a deeper insight into SAGDFN's performance, we perform visualizations on METR-LA and CARPARK1918 datasets to compare the prediction results of our SAGDFN model with the corresponding ground truth values.  As shown in Figure \ref{fig:visual}, our SAGDFN model could effectively capture both the trend and seasonality in datasets of different scales. For instance, our SAGDFN model is able to capture both short-term traffic peaks/dips and long-term daily cyclic patterns. This demonstrates that our model has good robustness and versatility over datasets of different scales. When compared to the ground truths, the prediction results exhibit similar trends and seasonality while maintaining a smooth time series. This indicates that our SAGDFN model can resist real-world noise without overfitting to such noise. These results demonstrate the robustness and effectiveness of the SAGDFN model in handling both commonly used and large real-world datasets.
\section{Conclusion}\label{sec:conclusion}
In this paper, we introduce a novel method named \textbf{S}calable \textbf{A}daptive \textbf{G}raph \textbf{D}iffusion \textbf{F}orecasting \textbf{N}etwork (SAGDFN)  to effectively capture the intricate spatial-temporal correlation underlying the large time series datasets and deliver state-of-the-art performance on multivariate time series forecasting tasks. In particular, SAGDFN utilizes the Significant Neighbors Sampling module to learn a set of most significant nodes with abundant information, and the Sparse Spatial Multi-Head Attention module to effectively capture the slim spatial correlations among the time series. Furthermore, the Encoder-Decoder-based Forecasting framework is employed to jointly model the multivariate time series and produce accurate predictions. Extensive experiments on four real-world datasets demonstrate that SAGDFN consistently outperforms all baselines on both commonly used and large datasets.
\section{Acknowledgement}
This study is supported under the RIE2020 Industry Alignment Fund – Industry Collaboration Projects (IAF-ICP)  Funding Initiative, as well as cash and in-kind contribution from Singapore Telecommunications Limited (Singtel), through Singtel Cognitive and Artificial Intelligence Lab for Enterprises (SCALE@NTU) and the National Research Foundation, Singapore under its AI Singapore Programme (AISG Award No: AISG2-TC-2021-001). Xiucheng Li is supported in part by the National Natural Science Foundation of China under Grant No. 62206074, Shenzhen College Stability Support Plan under Grant No. GXWD20220811173233001.

\normalem\bibliography{sample}
\bibliographystyle{IEEEtran}

\end{document}